\documentclass{article}

\usepackage{authblk}
\usepackage{booktabs}
\usepackage{multirow}
\usepackage[letterpaper,top=2cm,bottom=2cm,left=2.5cm,right=2.5cm,marginparwidth=1.75cm]{geometry}

\usepackage{amsmath,amssymb, bm}%, color}

\usepackage{graphicx}
\usepackage[colorlinks=true, allcolors=blue]{hyperref}
\usepackage{xcolor}
\newcommand{\fix}{blue} 

\title{%\textcolor{red}{(Confidential Information)}\\
Sparse-firing regularization methods for spiking neural networks with time-to-first spike coding }
\author[1,2]{Yusuke Sakemi}
\author[3]{Kakei Yamamoto}
\author[4]{Takeo Hosomi} 
\author[1,2]{Kazuyuki Aihara}

\affil[1]{Research Center for Mathematical Engineering, Chiba Institute of Technology, Narashino, Japan}
\affil[2]{International Research Center for Neurointelligence (WPI-IRCN), The University of Tokyo, Tokyo, Japan}
\affil[3]{Center for Advanced Intelligence Project, RIKEN, Tokyo, Japan}
\affil[4]{NEC Corporation, Kawasaki, Japan}
\begin{document}
\maketitle

\begin{abstract}
The training of multilayer spiking neural networks (SNNs) using the error backpropagation algorithm has made significant progress in recent years. Among the various training schemes, the error backpropagation method that directly uses the firing time of neurons has attracted considerable attention because it can realize ideal temporal coding. 
This method uses time-to-first spike (TTFS) coding, in which each neuron fires at most once, and this restriction on the number of firings enables information to be processed at a very low firing frequency.
This low firing frequency increases the energy efficiency of information processing in SNNs, which is important not only because of its similarity with information processing in the brain, but also from an engineering point of view. 
However, only an upper limit has been provided for TTFS-coded SNNs, and the information-processing capability of SNNs at lower firing frequencies has not been fully investigated.  
In this paper, we propose two spike timing-based sparse-firing (SSR) regularization methods to further reduce the firing frequency of TTFS-coded SNNs. 
The first is the membrane potential-aware SSR (M-SSR) method, which has been derived as an extreme form of the loss function of the membrane potential value. 
The second is the firing condition-aware SSR (F-SSR) method, which is a regularization function obtained from the firing conditions. 
Both methods are characterized by the fact that they only require information about the firing timing and associated weights. 
The effects of these regularization methods were investigated on the MNIST, Fashion-MNIST, and CIFAR-10 datasets using multilayer perceptron networks and convolutional neural network structures.
\end{abstract}

\section*{Introduction}

Spiking neural networks (SNNs) can process information in the form of spikes in a manner similar to the way information is processed in the brain.    
SNNs are thereby expected to be able to achieve both high computational functionality and energy efficiency \cite{Roy2019towards}. 
The spikes are represented as all-or-none binary values, 
and how information is represented by spikes is closely related to the information-processing mechanism in SNNs.
The spike-based information representation methods are divided into two major categories, rate coding and temporal coding \cite{Guo2021neural, Auge2021survey}. 
In rate coding, information is contained in the average number of spikes generated by a neuron. 
In this case, the firing frequency can take approximately continuous values as a function of the input intensities; therefore, the resulting SNNs can be treated as differentiable models similar to an artificial neural network (ANN). 
Using rate coding, ANNs can be converted to SNNs, and the high learning  ability of ANNs has been successfully transferred to SNNs \cite{Diehl2015fast,Ruechauer2017conversion, Kim2020spiking}. 
However, when rate coding is used, information processing in the SNNs is just an approximation of that in ANNs.   
Furthermore, the precise approximation of an ANN requires many spikes, which reduces energy efficiency when implemented in neuromorphic hardware \cite{Davies2018loihi}.  
It has been experimentally shown that physiologically, neurons in certain brain regions or specific neuron types exhibit extremely sparse firing characteristics \cite{Barth2012experimental}, and it is thought that temporal coding using not only the firing frequency but also the firing time is realized in at least some brain regions \cite{Fujiidynamical1996, Gollisch2008rapid,Portelli2016rank, Jaramillo2017phase}. 
Therefore, to achieve brain-like high-capacity, energy-efficient information processing capabilities in SNNs, it is important to use  temporal coding that also considers spike timing information.

% temporal coding and direct training of SNN
Because in temporal coding, the precise timing of spikes carries information, it is necessary to train the SNNs directly instead of using a converted ANN.  
In recent years, by incorporating deep learning techniques, it has become possible to directly train SNNs using the backpropagation algorithm \cite{Tavanaei2019Deep, Pfeiffer2018Deep, Dampfhoffer2023backpropagation, Eshraghian2023training}. 
Among the various methods proposed, methods that focus on the displacement of the membrane potential and those that focus on the displacement of the spike time have attracted particular attention because of their high learning performance. 
In membrane potential displacement methods, the derivative of the output spike with respect to the membrane potential is almost always zero because the spike is a binary value.  
However, it is possible to approximate this derivative using a surrogate function \cite{Neftci2019surrogate}. 
This method has been proposed in various forms by various groups \cite{Neftci2017event, Huh2018gradient, Zenke2018super, Wu2018spatio}. 
It has been proven to be very flexible, works with various surrogate functions \cite{Zenke2021remarkable}, and can be used to efficiently train recurrent neural network structures \cite{Yin2021accurate, Bellec2020solution}.
Recently, it has become possible to train relatively large models \cite{Kim2021revisiting}. 
However, with few exceptions \cite{Cramer2022surrogate,Yan2022backpropagation}, the neurons exhibit high firing frequencies, and it is debatable whether the timing information is efficiently utilized.

% about timing-based learning method
A timing-based learning method is a method that focuses on the displacement of the spike time \cite{Bohte2002error}. 
%Since This learning method exactly compute the gradient with respect to the spike time, an ideal form of temporal coding can be achieved. 
The coding most commonly used in this learning method is time-to-first-spike (TTFS) coding, which has the property that each neuron fires at most once \cite{Thorpe2001spike, Bonilla2022analyzing}. 
Because the information is contained in the timing of a single spike and the gradient is computed directly using the spike timing, this coding is expected to realize an ideal temporal coding.
The high learning performance of this method has been shown in various neuron models \cite{Mostafa2018supervised, Kheradpisheh2019s4nn, Comsa2021temporal, Sakemi2023supervised, Sakemi2021effects, Zhang2021rectified, Goltz2021fast}.
Hardware implementation efforts are also underway to achieve high power efficiency by taking advantage of its sparsity characteristics \cite{Goltz2021fast, Oh2022neuron}.
However, the constraint of firing at most once per neuron in TTFS coding may not be sufficiently sparse in some situations. 
For example, in the brain, there are many neurons that hardly fire at all \cite{Barth2012experimental}.
Furthermore, in an extremely power-limited environment such as edge AI \cite{Murshed2021machine}, a sparse firing pattern that goes beyond the constraint of one firing per neuron is desirable.

In this paper, we propose two methods to further improve the sparse firing property of TTFS-coded SNNs. 
One method is derived from the loss in the value of the membrane potential, and the other is derived from the firing conditions. 
Both methods are characterized by the fact that they only require information about the firing timing and the weights associated with it, as is the case in timing-based learning. 
In the following, we describe the two methods and show experimentally how  they suppress firing effectively on the MNIST, Fashion-MNIST, and CIFAR-10 datasets.

\section*{Results}

\subsection*{Spike timing-based sparse-firing regularization methods}

\begin{figure*}%[tbhp]
\centering
\includegraphics[clip, width=16cm]{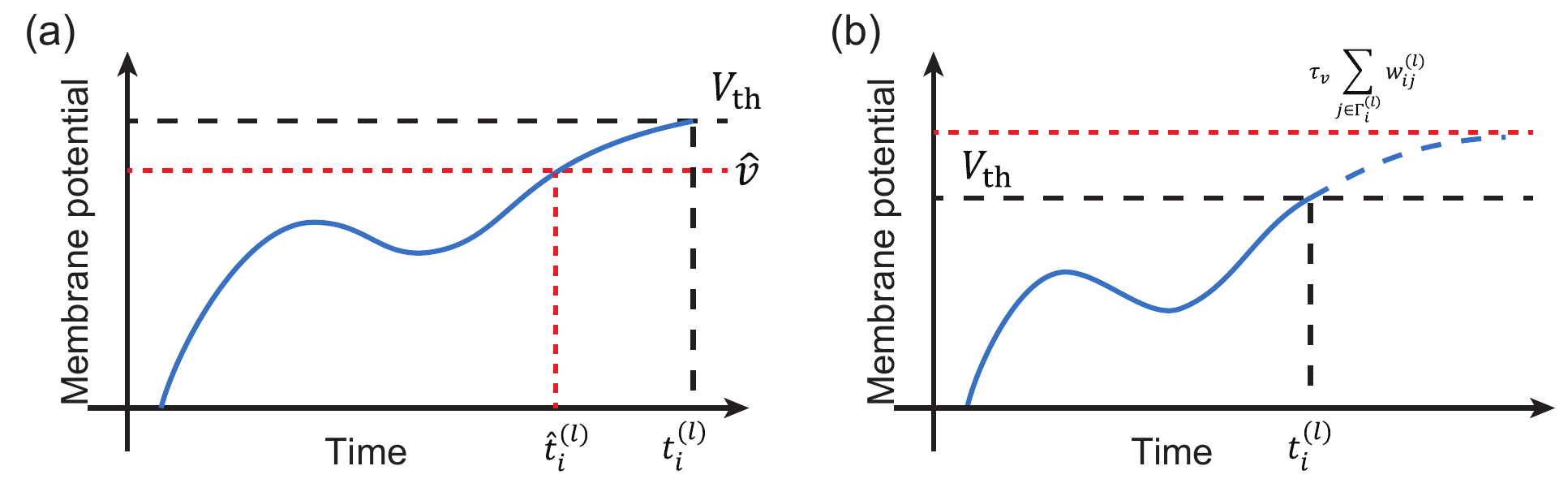}
\caption{{\bf Derivation of the two SSR methods}. (a)  M-SSR is derived from the membrane potential loss, where the loss occurs when the membrane potential is larger than $\hat{v}$. 
Assuming that a neuron fires at $t_i^{(l)}$ and $\hat{v}$ is sufficiently large, the loss occurs only in the time interval $[\hat{t}_i^{(l)}, t_i^{(l)}]$, 
where $\hat{t}_i^{(l)}$ is the time at which the membrane potential equals $\hat{v}$. 
M-SSR is obtained by setting $\hat{v}\rightarrow V_\text{th}$.
(b) F-SSR is derived from the firing condition. 
Assuming that no input spike is accepted after the neuron's firing time $t_i^{(l)}$, the membrane potential asymptotes to $\tau _v \sum _{j\in \Gamma_i^{(l)}}w_{ij}^{(l)}$ at $t\rightarrow \infty$. 
The F-SSR is derived by formulating the loss such that this asymptotic membrane potential becomes small. 
}
\label{fig:Vmem_loss}
\end{figure*}

We first summarize the proposed spike timing-based sparse-firing regularization (SSR) methods. 
SSR methods are characterized by the fact that they only require information about the firing timing and the weights associated with it, as is the case in ordinary timing-based learning. 
In this study, we propose two SSR method variants: 
membrane-potential-aware SSR (M-SSR) and  firing-condition-aware SSR (F-SSR). 
In both cases, we add a new regularization term to the cost function used in supervised learning to suppress the firing. 
Moreover, firing events can be suppressed by adding a new regularization term to the cost function used in supervised learning.
M-SSR is based on the idea of reducing the value of the membrane potential, which is realized by adding the membrane potential loss $V$ as a regularization term to the cost function. 
F-SSR is based on the idea of breaking the firing conditions, which is realized by adding the firing condition loss $Q$ as a regularization term to the cost function. 
Figure \ref{fig:Vmem_loss} shows the outline of each method.  
For simplicity, this paper adopts commonly used leaky integrate-and-fire (LIF) neuron models. 
The LIF neuron model has as parameters the time constant of the membrane potential $\tau _v$ and the time constant of the synaptic current $\tau _I$ (see Method).
Extensions to other neuron models are straightforward.  

First, we explain M-SSR, which is based on the idea of reducing the membrane potential value.  
The membrane potential loss $V$ is defined as 
\begin{flalign}
V &= \sum _l \xi ^l \sum _i V_i^{(l)}, \label{eq:Vmem_loss}\\
V_i^{(l)} &= \frac{1}{V_\text{th}-\hat{v}}\int _0 ^T dt \left( v_i^{(l)}(t) - \hat{v} \right)  \theta \left( v_i^{(l)}(t) - \hat{v} \right) \theta \left( t_i^{(l)} - t \right),  \label{eq:Vmem_integration}
\end{flalign}
where $V_i^{(l)}$ is the loss relating to the membrane potential trajectory of the $l$th-layer neuron $i$ and $\xi(>0)$ is the hyperparameter for leveling the sparsity in each layer.
In addition, $V_\text{th}$ is the firing threshold, and 
$T$ is a parameter specifying the time interval during which firing is suppressed. 
Figure \ref{fig:Vmem_loss} (a) shows the loss associated with the membrane potential trajectory of a neuron. 
Note that $\hat{v}$ is sufficiently large and there is only one point $\hat{t}_i^{(l)}$ at which the membrane potential equals $\hat{v}$. 
In this case, the loss is nonzero only during $[\hat{t}_i^{(l)}, t_i^{(l)}]$. 
To perform integration in Eq. (\ref{eq:Vmem_integration}), we need information about the value of the membrane potential at each time step.
However, by setting $\hat{v}\rightarrow V_\text{th}$, we can obtain $\hat{t}_i^{(l)} \rightarrow t_i^{( l)}$. This avoids the integral calculation, and Eq.  (\ref{eq:Vmem_integration}) can be solved analytically.
When computing the gradient of this integral, it is important to fix the integration range $[\hat{t}, t_i^{(l)}]$. 
If we do not treat it as a fixed value, the more rapidly the membrane potential rises, the smaller the membrane potential loss in Eq. (\ref{eq:Vmem_integration}) becomes, and thus the firing is not effectively suppressed.  
Finally, we obtain the following M-SSR: 
\begin{flalign}
V_i^{(l)} =
\begin{cases}
\frac{1}{\color{\fix} \sum _{j\in \Gamma _i^{(l)}} w_{ij}^{(l)}}\left[ {\color{\fix}t_i^{(l)}} \sum _{j\in \Gamma _i^{(l)}} w_{ij}^{(l)} - \sum _{j\in \Gamma _i^{(l)}} w_{ij}t_j^{(l-1)} \right], \text{ for }\tau_v = \tau_I =  \infty,  \\
\tau \frac{ 1   }{{\color{\fix}{\sum _{j\in \Gamma _i^{(l)}} w_{ij}^{(l)}}} - V_\text{th}\tau ^{-1}} \left[ \sum_{j\in \Gamma _i^{(l)}}  w_{ij}^{(l)} - \exp \left(- \frac{{\color{\fix}{t_i^{(l)}}}}{\tau}\right) a_i^{(l)}   \right], \text{ for } \tau_v = \infty, \tau_I = \tau, \\
2\tau {\color{\fix}{\alpha _i^{(l)}}} \left[ \exp \left(-\frac{{\color{\fix}{t_i}}}{2\tau}\right) b_i^{(l)}  - \exp \left(-\frac{{\color{\fix}{t_i}}}{\tau}\right)  a_i^{(l)} \right], \text{ for } \tau_v = 2\tau, \tau_I = \tau.
\end{cases} \label{eq:M_SSR}
\end{flalign}
Note that in the above equations, the gradients are not calculated for the variables shown in blue (they are considered to be constants in the gradient calculations). 
This corresponds to fixing the integration range $[\hat{t}, t_i^{(l)}]$. 
Constant terms not involved in the learning are excluded. 
$\Gamma _i^{(l)}$ denotes the index set of spikes that have been input to the $l$th-layer neuron $i$ up to firing time $t_i^{(l)}$. 
In Eq. (\ref{eq:M_SSR}), the following variables are defined
\begin{flalign}
a_i^{(l)}&= \sum _{j\in \Gamma _i^{(l)}} w_{ij}^{(l)} \exp \left( \frac{t_j^{(l-1)}}{\tau} \right),~ 
b_i^{(l)}= \sum _{j\in \Gamma _i^{(l)}} w_{ij}^{(l)} \exp \left( \frac{t_j^{(l-1)}}{2\tau} \right), \\
\alpha_i^{(l)} &=\frac{2a_i^{(l)}}{\left(b_i^{(l)} + \sqrt{\left(b_i^{(l)}\right)^2-2a_i^{(l)}\tau^{-1}V_\text{th}}\right)\left(\sqrt{\left(b_i^{(l)}\right)^2 - 2a_i^{(l)}\tau^{-1}V_\text{th}}\right)}.
\end{flalign}
Appendix \ref{ss:derivation_appendix} provides a detailed derivation.
In addition, Appendix \ref{ss:convergence} discusses the consistency of the M-SSR gradient (Eq. (\ref{eq:M_SSR})) with that of the integral-form loss (Eq. (\ref{eq:Vmem_integration})) when $\hat{v}\rightarrow V_\text{th}$.

Next, we explain F-SSR, a method that suppresses firing based on the firing conditions. 
From the non-leaky integrate-and-fire neuron model $(\tau _v = \infty,~\tau _I = \tau)$, we obtain the following firing conditions: 
\begin{flalign}
\text{firing condition}_i^{(l)} := \sum_{j\in \Gamma _i^{(l)}} w_{ij}^{(l)} \ge V_\text{th} \tau^{-1}. 
\end{flalign}
Because the firing will be suppressed if this firing condition is not satisfied, we define the F-SSR term $Q$ as follows: 
\begin{flalign}
Q &= \sum _{l} \xi^l \sum _i Q_i^{(l)} \label{eq:F_loss} \\
Q_i^{(l)} &=  \begin{cases}
\sum _{j\in \Gamma _i^{(l)}} w_{ij}^{(l)}, ~\text{if } t_i^{(l)} < T \\
0,~\text{otherwise.} \\
\end{cases}\label{eq:F_SSR}
\end{flalign}
We note that $V_i^{(l)}=Q_i^{(l)}=0$ if the neuron does not fire.

\subsection*{Numerical simulations}

We trained several SNNs on the MNIST dataset \cite{Lecun1998gradient}, Fashion-MNIST dataset \cite{Xiao2017fashion}, and CIFAR-10 dataset \cite{CIFAR10} to investigate the effect of SSR on suppressing firing. 
In the experiment, in addition to the multilayer perceptron (MLP) structure, a convolutional neural network (CNN) structure was used. 
The image data in the dataset were converted to input spikes, where the intensity of each pixel is converted to the input time of each spike (see Method).
We define sparsity as the average number of spikes per neuron per input data in a time window $[0,t^\text{ref}]$, where $t^\text{ref}$ is the reference time of the output layer firing time (see Method).
In addition, we set $T=t^\text{ref}$ in Eqs. (\ref{eq:Vmem_integration}) and (\ref{eq:F_SSR}), and set the firing threshold $V_\text{th}$ to 1. 
When the integration form Eq. (\ref{eq:Vmem_integration}) was used, the integration was approximated by dividing the integral by the time width $\Delta t$.

\begin{figure*}%[h]
\begin{center}
\includegraphics[clip,width=16cm]{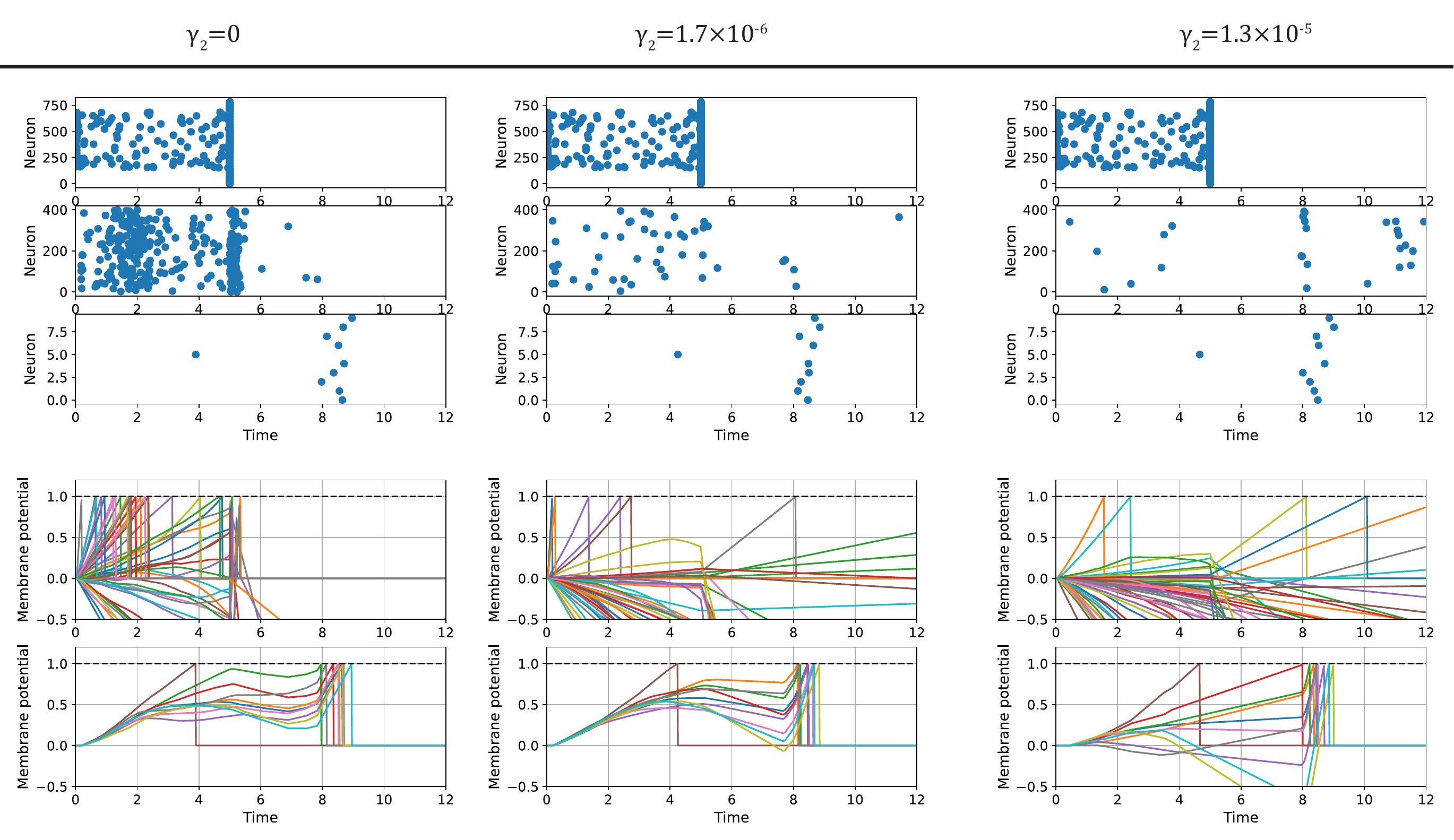}
\caption{{\bf Typical results with M-SSR.} 
We trained SNNs with one hidden layer (784-400-10) on the MNIST dataset with various values of M-SSR strengths $\gamma_2$.  
The upper figures  present raster plots for a given input data, showing the results of the input layer (top), hidden layer (middle), and output layer (bottom). 
The lower figures show the time evolution of the membrane potentials when given the same input data, with the results for the hidden layer (top) and the output layer (bottom). 
In the panels displaying the membrane potentials in the hidden layer, only 50 neurons are shown.
In all cases, the following hyperparameters were used: 
 $\tau_v = \tau_I = \infty,~t^\text{ref}=8,~\gamma_1 = 10^{-4}, \gamma_3=0, \eta = 10^{-4}$, and $\tau _\text{soft}=0.9$. 
}
\label{fig:typical_results}
\end{center}
\end{figure*}

Figure \ref{fig:typical_results} shows the learning results of SNNs with one hidden layer (784-400-10) trained on the MNIST dataset with various M-SSR  strengths $\gamma _2$ (see Method). 
Note that all output layer neurons were required to fire because the loss function was defined by the spike timing of the neurons in the output layer.   
Therefore, sparse firing regularization was applied only to the hidden layers. 
The upper figures show the raster plots of the firing distribution of each layer for a given input data, and the lower figures  show the time evolution of the membrane potentials of each layer. 
As the strength of M-SSR was increased, the number of neurons that fired tended to decrease, and it can be seen that most of the hidden layer neurons stopped firing when $\gamma _2=1.3\times 10^{-5}$.  
By contrast, the firing distribution of the output layer did not change significantly with respect to M-SSR strength.
The neurons corresponding to the correct index fired the earliest $(t\sim 4)$, and the other neurons fired later $(t\sim 8)$.  
The membrane potentials of the hidden layer neurons were suppressed as the regularization strength increased, whereas the output layer solved the task using fewer spikes from the hidden layer. 
This indicates that M-SSR regularization can suppress the firing of the hidden layer without significantly compromising recognition task performance. 

\begin{figure*}%[h]
\begin{center}
\includegraphics[clip,width=\textwidth]{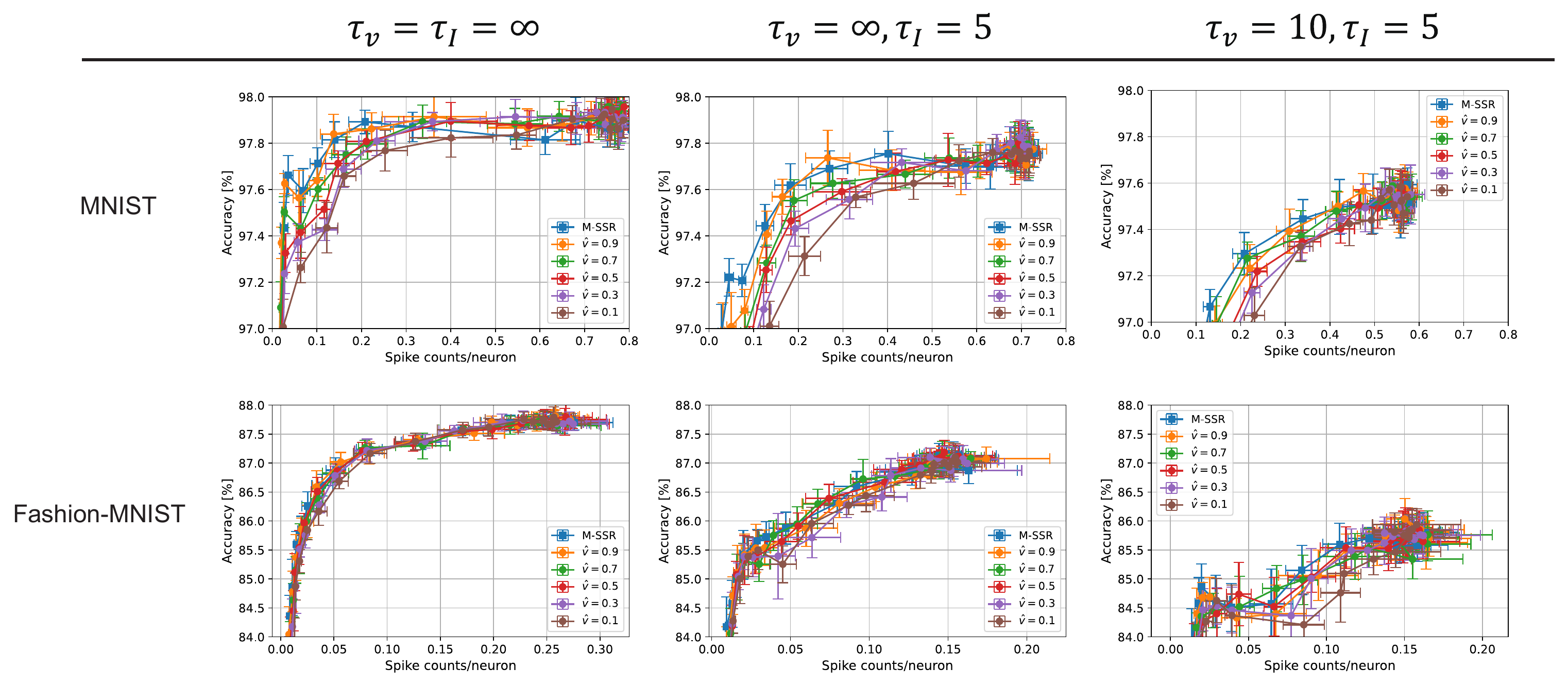}
\caption{{\bf Sparsity--accuracy tradeoff for different regularization forms.} 
We evaluated the integral-form regularization (Eq. (\ref{eq:Vmem_integration})) and M-SSR (Eq. (\ref{eq:M_SSR})) in a 784-400-10  SNN based on various neuron models. 
The tradeoff in sparsity--accuracy is shown when the regularization strength $\gamma _2$ was varied from $0$ to $10^{-4}$.  
The standard deviations were obtained over 10 trials. 
The hyperparameters were $t^\text{ref}=8, \gamma_1 = 10^{-4}, \gamma_3=0, \eta = 10^{-4}$, and $\tau_\text{soft}=0.9$. 
For the integral-form regularization, we used various values of $\hat{v}$, and we set $\Delta t = t^\text{ref}/1000$. 
}
\label{fig:sparsity_analysis}
\end{center}
\end{figure*}

Figure \ref{fig:sparsity_analysis} shows the sparsity--accuracy tradeoff results obtained when using the integral-form regularization (Eq. (\ref{eq:Vmem_integration})) and M-SSR (Eq. (\ref{eq:M_SSR})).  
We trained SNNs with a single hidden layer (784-400-10) using various regularization strengths. 
The standard deviations were obtained over 10 trials. 
The upper figures show the results for the MNIST dataset, and the lower figures show the results for the Fashion-MNIST dataset. 
Results are also shown for different neuron models $(\tau _v, \tau _I)$.
For the MNIST dataset, the tradeoff curves show that a larger $\hat{v}$ led to a better tradeoff for all neuron models, and the best tradeoff was obtained by M-SSR (corresponding to $\hat{v}=1$). 
Similar results were obtained for Fashion-MNIST, although the advantage was not as pronounced as it was for MNIST. 
This result demonstrates that the integral-form regularization (Eq. (\ref{eq:Vmem_integration})) smoothly transitioned to the limit form (M-SSR, Eq. (\ref{eq:M_SSR})).
Furthermore, taking the limit of $\hat{v}\rightarrow 1$, the tradeoff between sparsity and accuracy was improved. 
In addition, good sparsity--accuracy tradeoff properties were obtained for various neuron models ($\tau_v$ and $\tau_I$), suggesting that M-SSR is preferable to the integral-form regularization for TTFS-coded SNNs. 

\begin{figure*}%[h]
\begin{center}
\includegraphics[clip,width=\textwidth]{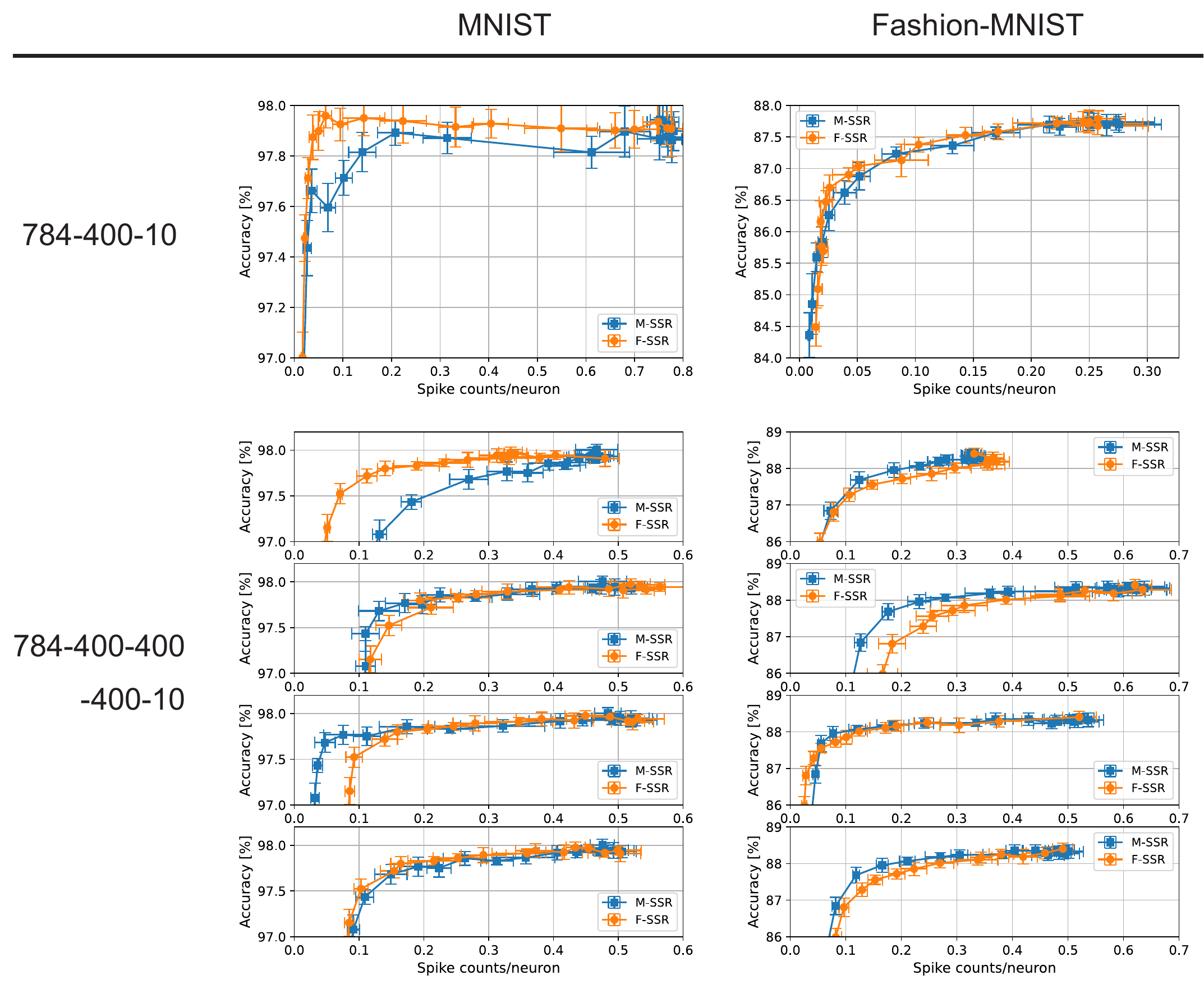}
\caption{{\bf Comparison of SSR methods on the MNIST and Fashion-MNIST datasets.} The top figures show the sparsity--accuracy tradeoffs for the 2-layer SNNs (784-400-10), whereas the bottom figures show the sparsity--accuracy tradeoffs for the 4-layer SNNs (784-400-400-400-10). In the bottom figures, each panel represents the accuracy--sparsity tradeoff for the first, second, and third hidden layers, from the top. The bottom panel  presents the sparsity--accuracy tradeoffs for the sparsity averaged over the three hidden layers. 
The standard deviations were obtained over 10 trials. 
We used the following hyperparameters: $t^\text{ref}=8, \gamma_1 = 10^{-4}, \eta = 10^{-4}$, and $\tau_\text{soft}=0.9$ for the 2-layer SNNs, $t^\text{ref}=9, \gamma_1 = 10^{-4}, \eta = 10^{-4}$, and $\tau_\text{soft}=0.9$ for the 4-layer SNNs. For the MNIST dataset, we set $\xi$ to 6 for M-SSR, and 1 for F-SSR.  For the Fashion-MNIST dataset, we set $\xi$ to 6 for M-SSR, and 4 for F-SSR. 
}
\label{fig:SSRs}
\end{center}
\end{figure*}

Figure \ref{fig:SSRs} compares the results of the two proposed SSRs, M-SSR (Eq. (\ref{eq:M_SSR})) and F-SSR (Eq. (\ref{eq:F_SSR})).  
The results for an SNN with one hidden layer (784-400-10) are shown in the top figures. 
On the MNIST benchmark, F-SSR obtained a better sparsity--accuracy tradeoff than M-SSR, 
whereas on the Fashion-MNIST benchmark, both F-SSR and M-SSR yielded a similar sparsity--accuracy tradeoff. 
The results for an SNN with three hidden layers are shown in the lower figures. 
For each SSR method for each dataset, the value of $\xi$ (Eq. (\ref{eq:Vmem_loss})) was set to obtain the best tradeoff averaged over the whole layer (except for the output layer). 
See Appendix \ref{ss:magnification} for the tradeoff properties for various values of $\xi$.
For an SNN with three hidden layers, M-SSR and F-SSR showed similar sparsity--accuracy tradeoff characteristics, but the optimal value of $\xi$ differed significantly in M-SSR  and F-SSR. 
For the MNIST dataset, the optimal value was $\xi=6$ for M-SSR and $\xi=1$ for F-SSR,  
whereas for the Fashion-MNIST dataset, the optimal value was  $\xi=6$ for M-SSR and $\xi=4$ for F-SSR. 
This difference may be due to the characteristics of the regularization function. 
In M-SSR (Eq. (\ref{eq:M_SSR})), the error that occurred in the $l$-th layer propagates back to the previous layer via spike timing $t_j^{(l-1)}$. 
In this case, if the weights of neurons from $t_j^{(l-1)}$ to the $l$th layer were positive overall, $t_j^{(l-1)}$ increased during training, and consequently the $l-1$th layer also became more sparse. 
Similarly, the $l-2$th layer was expected to become sparse. 
Therefore, to counteract this effect, a relatively large value of $\xi$ was optimal. 
By contrast, in the case of F-SSR (Eq. (\ref{eq:F_SSR})), the losses that occurred in the $l$th layer do not propagate back to previous layers. 
Therefore, a relatively small value of $\xi$ was optimal.

\begin{table} 
\begin{center}
\caption{{\bf Convolutional architectures used for each dataset.} 
``Conv($a$,$b$)'' represents a convolutional layer with a kernel size of $a\times a$, number of output channels $b$, and stride 1. 
``Pool'' represents a pooling layer with kernel size of $2\times2$, stride 2.
The padding of the convolutional layer was set to 0 for MNIST and set to 1 for Fashion-MNIST and CIFAR-10. }
\begin{tabular}{ll }
Dataset & Network  \\ \hline  
MNIST & Conv(5, 6)-Pool-Conv(5, 16)-Pool-400-400-10 \\
Fashion-MNIST &  Conv(5, 6)-Pool-Conv(5, 16)-Pool-400-400-10 \\
CIFAR-10 & Conv(3, 24)-Pool-Conv(3, 48)-Pool-Conv(3, 96)-Pool-600-10
\end{tabular}
\label{tab:CNN} 
\end{center}
\end{table}

\begin{figure*}%[h]
\begin{center}
\includegraphics[clip,width=\textwidth]{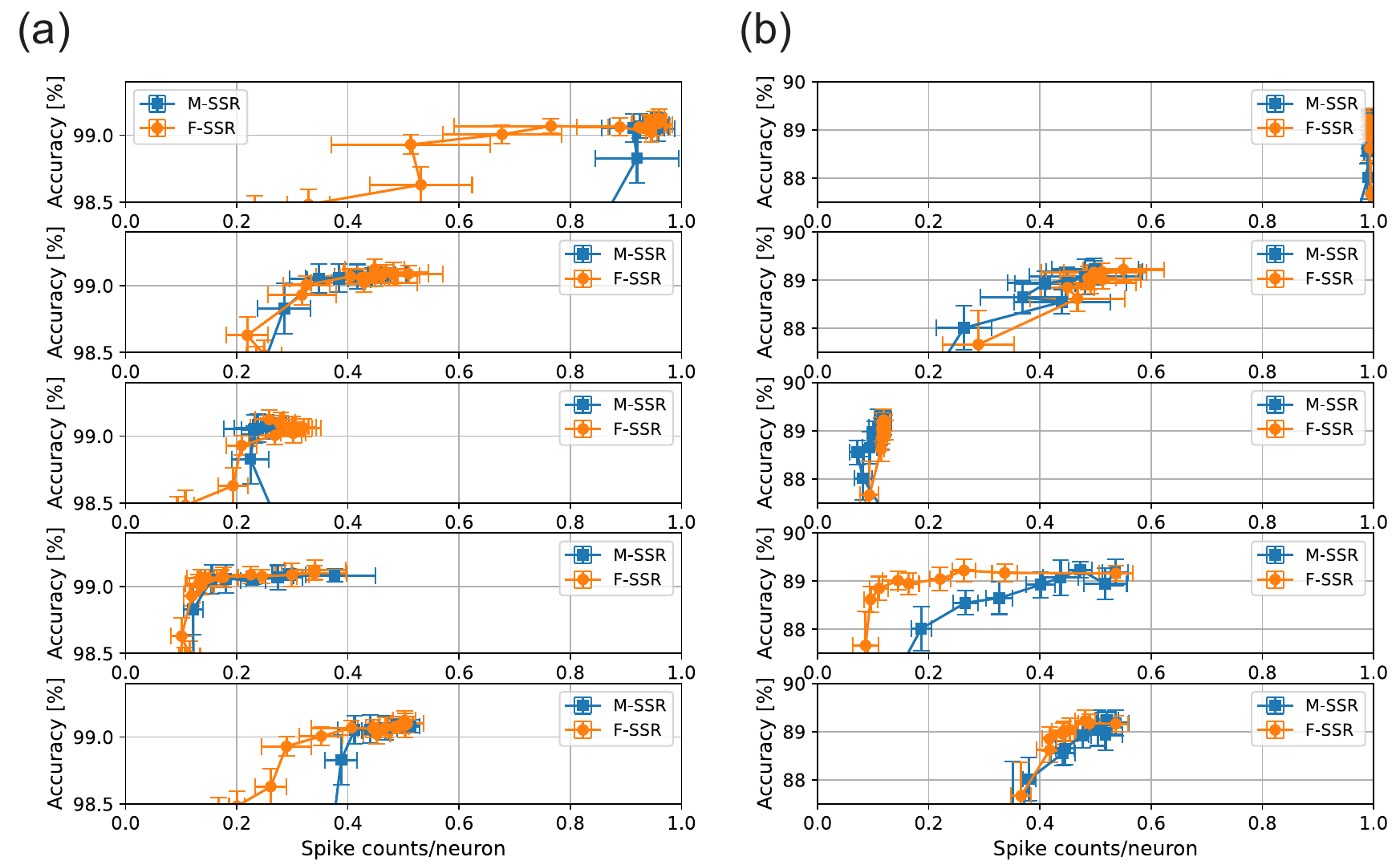}
\caption{{\bf Effects of SSR with a CNN-architecture on the MNIST dataset (a) and the Fashion-MNIST dataset (b).} From top to bottom, the panels represent the sparsity--accuracy tradeoffs for the first convolutional layer, second convolutional layer, first fully connected layer, second fully connected layer, and whole network (not including pooling layers). 
The standard deviations were obtained over 10 trials. 
We used the following hyperparameters: $t^\text{ref}=16, \gamma_1 = 10^{-4}, \eta = 10^{-4},$ and $\tau_\text{soft}=0.9$. For the MNIST dataset, we set $\xi$ to 6 for M-SSR, and 2 for F-SSR.  For the Fashion-MNIST dataset, we set $\xi$ to 6 for M-SSR, and 4 for F-SSR.
} 
\label{fig:LeNet}
\end{center}
\end{figure*}

Next, we applied the SSR methods to spiking CNNs. 
Table \ref{tab:CNN} shows the network structure used for each dataset. 
Figure \ref{fig:LeNet} shows the effect of SSR regularization on MNIST and Fashion-MNIST. 
The overall sparsity--accuracy tradeoff for the CNN structure was worse than that for the MLP structure.
The SNNs with MLP structures reduced the average number of firings per neuron to about $0.1-0.2$ with almost no loss in accuracy, whereas the SNNs with convolutional structures only reduced the number of firings to about $0.4$. 
It can be seen that the first convolutional layers are not very sparse, with the exception of the results of F-SSR on the MNIST dataset. 
This is considered to be caused by the fact that it is difficult to force only a portion of the neurons to fire in a convolutional layer because of the weight-sharing property. 

\begin{figure}%[h]
\begin{center}
\includegraphics[clip,width=8cm]{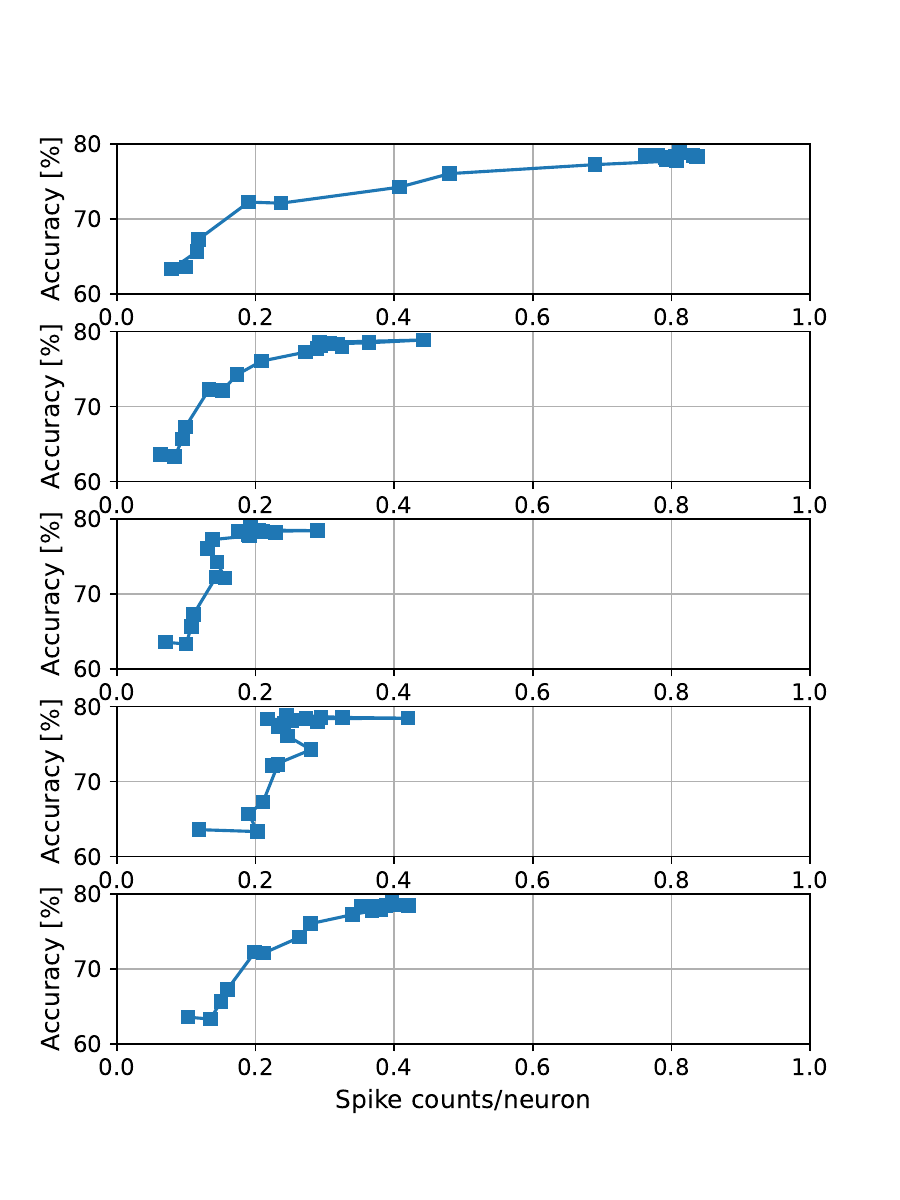}
\caption{{\bf Sparsity--accuracy tradeoff on the CIFAR-10 task.} The best case over 10 trials is plotted. 
From the top, the sparsity--accuracy tradeoffs for the first, second, and third convolutional layers, and then fully connected layer are represented, respectively. The sparsity—accuracy tradeoff for the whole network (not including pooling layers) is presented in the bottom panel. 
We inverted the effects of  F-SSR,  so the neurons tend to fire more frequently. } 
\label{fig:VGG}
\end{center}
\end{figure}

On the CIFAR-10 dataset, firing tended to be suppressed during training even when SSR was not applied, and we confirmed that neurons relating to some channels did not fire at all image locations and for all training data.  
To avoid this problem, we trained the SNNs to attempt to satisfy the firing conditions. 
This was achieved by making the regularization strength $\gamma_3$ of the F-SSR term negative (see Method).  
The results of learning by promoting firing are shown in Fig.\ref{fig:VGG}.
Because the variance over the 10 trials was large, we plotted the results for the trial with the highest test accuracy for better visualization.
Appendix \ref{ss:raw_plot} presents the results for all trials. 
By sacrificing sparsity, we were able to observe a noticeable improvement in performance.
We conducted an experiment incorporating M-SSR regularization, but the combination of promoting firing and sparse firing did not yield a favorable sparsity-accuracy tradeoff.

\section*{Discussion}

SNNs with TTFS coding can realize ideal temporal coding by constraining each neuron to fire at most once. 
Due to this mechanism, the SNNs with TTFS coding have high firing sparsity, and this approach has been applied in energy-efficient hardware implementations \cite{Goltz2021fast, Oh2022neuron}.  
To further improve this sparse firing characteristic, we developed the SSR methods. 
The two SSR methods were derived from two different perspectives.
The first one is M-SSR, which was derived by assigning a penalty each time the membrane potential exceeds a threshold $\hat{v}$ and taking the limit when the threshold equals the firing threshold. 
The other is F-SSR, which was obtained from the firing conditions of neurons. 
Both SSR methods are characterized by the fact that they do not require information about the membrane potential itself, only the firing time and associated weights. 
The sparsity--accuracy properties of these two methods were investigated using the MNIST, Fashion-MNIST, and CIFAR-10 datasets.
Interestingly, although some differences were observed depending on the datasets and the network structure, both F-SSR and M-SSR showed equally good sparsity--accuracy properties, even though the regularization methods were derived from different perspectives.
In particular, for the fully connected layer, it was found that the average number of firings for each neuron could be lowered to 0.1 to 0.2. 
From the experiments conducted in this study, it is difficult to determine which method is superior. We can at least conclude that F-SSR has the advantage of a somewhat smaller computational load than M-SSR due to its simpler formula.
To understand the difference between F-SSR and M-SSR, in addition to the sparsity--accuracy property, a detailed analysis of the changes in the firing characteristics and in the information processing mechanism associated with the sparse firing mechanism will be required in future. 

% 大規模CNNに向けた課題
For the CNN structures, we found the SSR methods had more difficulty suppressing the firing of neurons in the convolutional layer than in the fully connected layer.  
In CIFAR-10 in particular, we observed that firing is suppressed too much and learning becomes difficult even without SSR. 
This may be because it is difficult to flexibly decide whether the outputs belonging to a certain kernel should fire depending on the position because of the weight-sharing property. 
To prevent this, we found that, in CIFAR-10, learning performance can be improved by promoting firing. 
Similar firing promotion terms have been introduced in previous studies \cite{Mostafa2018supervised, Zhou2021temporal}.%, Zhang2020temporal}. 
In timing-based learning of large-scale CNN structures, one way to obtain better sparsity--accuracy properties is to utilize models that allow multiple neuron firings \cite{Yamamoto2022timing} combined with the SSR methods.

% Difference between SSR and previous methods
Previous studies have developed methods that suppress the firing of SNNs in the framework of the surrogate gradient method \cite{Cramer2022surrogate,Yan2022backpropagation}. 
They applied direct regularization to the spike variable $s(t)\in \{0, 1\}$ represented at each time step in the model to the time-discretized SNN. 
The gradient calculation is made possible using the surrogate function $\frac{d s(t)}{dv(t)}=\sigma (v(t))$ \cite{Neftci2019surrogate}. 
This method is closely related to the M-SSR proposed in this paper.
In the surrogate gradient method, the spike variables above are treated as a function of the membrane potential $s(t)=\int _{-\infty} ^{v(t)} \sigma(v') dv'$. 
In this sense, the idea is similar to the loss in Eq. (\ref{eq:Vmem_integration}), which integrates the membrane potential.
By contrast, M-SSR, unlike the previous method \cite{Cramer2022surrogate,Yan2022backpropagation}, can be transformed from the time-integration form to the timing form by setting $\hat{v}\rightarrow V_\text{th}$.
This may correspond to the fact that the learning method with the surrogate function can transition to a timing-like learning method by taking a limit  \cite{Huh2018gradient}.
Interestingly, as shown in Fig.\ref{fig:sparsity_analysis}, the sparsity--accuracy property improves as $\hat{v}$ gets closer to $V_\text{th}$. 
This suggests that timing-based sparse regularization is more effective in timing-based learning.  
We note that F-SSR is a regularization method using firing conditions, which is unique to timing-based learning.

% about SNN hardware and other sparse techniques
SNNs can operate efficiently on neuromorphic hardware \cite{Cramer2022surrogate, Moradi2018scalable}. 
Because the energy consumed by the spike transmission increases as the firing frequency increases, reducing the firing frequency is an important  issue in real-world applications. 
SNNs with TTFS coding are expected to provide significant power advantages in hardware implementation due to their extremely sparse firing \cite{Sakemi2023supervised,  Sakemi2021effects, Sakemi2022spiking}.
Several research groups have reported hardware implementations of such SNNs \cite{ Oh2022neuron,Goltz2021fast,Oh2020hardware}. 
The SSR methods are expected to further improve the energy efficiency of SNNs. 
Moreover, unlike the methods in \cite{Cramer2022surrogate,Yan2022backpropagation}, the SSR methods can calculate the gradient without observing the membrane potential, which may simplify the learning system on hardware. 
Finally, in addition to the reduction in the firing rate, the combination of binarized weights \cite{Kheradpisheh2022bs4nn} and pruned weights \cite{Faghihi2022synaptic, Yan2022backpropagation, Han2022adaptive} is expected to make the SNN model more suitable for hardware implementation.

\section*{Method}

\subsection*{SNN models}

In this study, we constructed a multilayer SNN using the following LIF neuron model: 
\begin{flalign}
\frac{d}{dt} v_i ^{(l)} (t) &= \frac{1}{\tau_v} v_i ^{(l)} (t) + I_{i}^{(l)} ,  \label{eq:diff_neuron}\\
\frac{d}{dt} I_i ^{(l)} (t) &=\frac{1}{\tau_I} I_i ^{(l)} (t)  +  \sum _{j=1} ^{N^{(l-1)}} w_{ij}^{(l)} \delta(t - t_j ^{(l-1)}), 
\end{flalign}
where $v_i^{(l)}$ is the membrane potential of neuron $i$ in the $l$th layer, $w_{ij}^{(l)}$ is the coupling strength from neuron $j$ in the $l-1$th layer to neuron $i$ in the $l$th layer, and $t_j^{(l-1)}$ is the firing time of neuron $j$ in the $l-1$th layer.
Furthermore, $\tau_v$ is the time constant of the membrane potential and $\tau_I$ is the time constant of the synaptic current. 
$N^{(l)}$ is the number of neurons that make up the $l$th layer. 
Neurons fire when the membrane potential reaches the firing threshold $V_\text{th}$ and generate spikes.
After firing, the membrane potential is fixed at $0$ and never fires again. 
The membrane potential of the model described in Eq. (\ref{eq:diff_neuron}) is analytically obtained as follows:
\begin{flalign}
v_i^{(l)}(t) &= \frac{\tau _v \tau_I}{\tau _v - \tau _I} \sum_{j=1}^{N^{(l-1)}} w_{ij}^{(l)}\kappa(t -t_j^{(l-1)}),  \label{eq:neuron}\\
\kappa(t) &= \theta(t) \left[ \exp \left(-\frac{t}{\tau _v}\right) - \exp \left(- \frac{t}{\tau _I}\right)\right], \\
\theta (t) &= \begin{cases}
0, \text{ for } t<0, \\
1, \text{ for } 0\le t. 
\end{cases}
\end{flalign}
The experiments in this study consider the three cases $(\tau_v, \tau_I) \in \{ (2\tau, \tau), (\infty, \tau)$, and $(\infty, \infty)\}$.
Note that the learning characteristics of SNNs with TTFS coding were investigated for the cases of $\tau_v=\infty, \tau_I = \infty$, \cite{Sakemi2023supervised}, $\tau_v=\infty$ \cite{Mostafa2018supervised}, and $\tau _v \neq \infty,\tau_I \neq \infty$ \cite{ Comsa2021temporal,Goltz2021fast}. 

The firing time in each case can be calculated from the condition $v_i^{(l)}(t_i^{(l)})=V_\text{th}$ as follows: 
\begin{flalign}
t_i^{(l)} = \begin{cases}
\frac{V_\text{th} + \sum _{j\in \Gamma_i^{(l)}}  w_{ij}^{(l)}t_j^{(l)}}{\sum _{j\in \Gamma_i^{(l)}}  w_{ij}^{(l)}}, \text{ for } (\tau_v, \tau_I)=(\infty, \infty),\\
\tau \ln\left[\frac{\sum _{j\in\Gamma_i^{(l)}} w_{ij}^{(l)} \exp \left( \frac{t_j^{(l-1)}}{\tau} \right)}{\sum _{j\in\Gamma_i^{(l)}} w_{ij}^{(l)} - V_\text{th}\tau^{-1}}\right], \text{ for }(\tau_v, \tau_I)=(\infty, \tau),\\
2\tau \ln\left[\tau\frac{b_i^{(l)} - \sqrt{(b_i^{(l)})^2 - 2a_i^{(l)}\tau^{-1}V_\text{th}}}{V_\text{th}} \right], \text{ for }(\tau_v, \tau_I)=(2\tau, \tau),
\end{cases}
\end{flalign}
where $\Gamma _i^{(l)}$ denotes the index set of spikes input to the $l$th layer neuron $i$ up to firing time $t_i^{(l)}$. 
We define the following variables: 
\begin{flalign}
a_i^{(l)}= \sum _{j\in\Gamma_i^{(l)}} w_{ij}^{(l)} \exp \left( \frac{t_j^{(l-1)}}{\tau} \right),~  b_i^{(l)}= \sum _{j\in\Gamma_i^{(l)}} w_{ij}^{(l)} \exp \left( \frac{t_j^{(l-1)}}{2\tau} \right).
\end{flalign}
A detailed derivation of the firing time in the case of $(\tau_v, \tau_I) = (\tau, 2\tau)$ is given in Appendix \ref{ss:alpha_synaptic}. 

\subsection*{Learning algorithms}
Supervised learning of the SNN was performed using the following cost function
\begin{flalign}
C  &= L(t^{(M)}) + \gamma_1 T(t^{(M)}) + \gamma_2 V + \gamma_3 Q, \label{eq:cost_function}\\
L  &= \sum _{i=1}^{N^{(M)}} \kappa _i \ln S_i, \\
S_i&= \frac{\exp\left(\frac{t_i^{(M)}}{\tau _\text{soft}}\right)}{\sum_{j=1}^{N^{(M)}} \exp \left(\frac{t_j^{(M)}}{\tau _\text{soft}}\right)}, \\
T  &= \sum _{i=1}^{N^{(M)}}\left(t_i^{(M)} - t^\text{ref}\right) ^2, 
\end{flalign}
where $M$ represents the output layer and $t^{(M)}=\left(t_1^{(M)}, t_2^{(M)}, \dots, t_{N^{(M)}}^{(M)}\right)$.
The value of the teacher label $\kappa_i$ is equal to one when the $i$th label is assigned and zero otherwise.  
Parameters $\gamma _1$, $\gamma _2$, and $\gamma_3$ are real numbers, and they respectively control the signiﬁcance of the temporal penalty term $T$, the membrane potential loss $V$ (Eq. (\ref{eq:Vmem_loss})), and the firing condition term $Q$ (Eq. (\ref{eq:F_loss})). 
Parameter $\tau _\text{soft}$ is a positive real number, which adjusts the softmax scaling. 
Learning was performed by minimizing this cost function using the gradient method with the Adam optimizer \cite{Kingma2014adam} at a learning rate of $\eta$.  
On the CIFAR-10 task, the coefficient $\gamma _3$ was set to a negative number to promote firing. 
In this case, the firing condition term $Q$ was modified as follows: 
\begin{flalign}
Q_i^{(l)} &=  \begin{cases}
\sum_{i=1}^{N^{(l-1)}} w_{ij}^{(l)}, ~\text{if not fired}, \\
0,~\text{otherwise.} \\
\end{cases}
\end{flalign}
We promoted the neurons to fire only when the corresponding neurons did not fire.

\subsection*{Dataset}

The MNIST, Fashon-MNIST, and CIFAR-10 datasets include 2-dimensional image data. 
In the MNIST and Fashion-MNIST datasets, each image has one channel, 
whereas in the CIFAR-10 dataset, the images have three channels. 
To process such image data, we first normalized the pixel intensity to $[0, 1]$. 
Then, we obtained an input spike as follows: 
\begin{flalign}
t_{ijk}^{(0)} = \tau _\text{in} (1-x_{ijk}),
\end{flalign}
where $x_{ijk}$ is the normalized pixel intensity, the first and second indices represent the coordinates of the pixel, and the third index represents the channel number. 
Here, $\tau _\text{in}$ is a positive constant. 
We set $\tau _\text{in}=5$ in all experiments.  
When spikes are input to a fully connected layer, the input tensors are reshaped into one-dimensional tensors.
For the CIFAR-10 dataset, 
to avoid the problem of the first hidden layer firing too early and ignoring later inputs, the number of channels was doubled as follows:
\begin{flalign}
x_{i,j,k} =  1 - x_{i,j,k-3}~(k=3,4,5).
\end{flalign}
Furthermore, we used data augmentation (horizontal flipping, rotation, and cropping) as in the previous study \cite{Zhou2021temporal}.

\section*{Acknowledgements}
This work was partially supported by 
JST PRESTO Grant Number JPMJPR22C5, 
NEC Corporation, 
SECOM Science and Technology Foundation, 
JST Moonshot R\&D Grant Number JPMJMS2021, 
AMED under Grant Number JP23dm0307009, 
the International Research Center for Neurointelligence (WPI-IRCN) at The University of Tokyo Institutes for Advanced Study (UTIAS), 
JSPS KAKENHI Grant Number JP20H05921. 
Computational resource of AI Bridging Cloud Infrastructure (ABCI) provided by National Institute of Advanced Industrial Science and Technology (AIST) was used.

\appendix

\section*{Appendices}

\section{Derivation of M-SSR (Eq. (\ref{eq:M_SSR}))} \label{ss:derivation_appendix}

%Appendixが別資料として提出しなければならないときは, Fig. 1をこっちに再掲する. 
In this section, we present a detailed derivation of M-SSR (Eq.  (\ref{eq:M_SSR})). 
If $\hat{v}$ is sufficiently close to $V_\text{th}$, $\hat{t}_i^{(l)}$, the time at which the membrane potential is $\hat{v}$ can be assumed to be a single point (see Fig. \ref{fig:Vmem_loss} (a)). 
Moreover, we can assume that the number of input spikes can be assumed to be constant $\left(\Gamma _i^{(l)}\right)$ in the time interval $[\hat{t}_i^{(l)}, t_i ^{(l)}]$. 
From the above, the integral (Eq. (\ref{eq:Vmem_integration})) can be transformed as follows
\begin{flalign}
V_i^{(l)} &=  \frac{1}{V_\text{th}-\hat{v}} \int_0 ^T dt   \left( v_i^{(l)}(t) - \hat{v} \right)  \theta \left( v_i^{(l)}(t) - \hat{v} \right) \theta \left( t_i - t \right), \\
&= \frac{1}{V_\text{th}-\hat{v}} \int_{\hat{t}_i^{(l)}} ^{t_i^{(l)}} dt \left( v_i^{(l)}(t) - \hat{v} \right).  
\end{flalign}
Note that $V_i^{(l)}=0$ if the neuron does not fire. 
Importantly, when computing the gradient of this integral, the integration range $[\hat{t}, t_i^{(l)}]$ should be fixed.   
This is because if they are not treated as fixed values, the SNNs learn to make the membrane potential rise rapidly, thus the effect of firing suppression cannot be obtained. 
We can omit the constant term $\hat{v}(t_i^{(l)} - \hat{t}_i^{(l)})$, which is not involved in the learning process, and we only need to calculate the following:  
\begin{flalign}
V_i = \frac{1}{V_\text{th}-\hat{v}} \int_{\hat{t}_i^{(l)}} ^{t_i^{(l)}}   v_i^{(l)}(t) dt.
\end{flalign}

In the following, the limit of the above integral $(\hat{v}\rightarrow V_\text{th})$ is calculated for various neuron models ($\left(\tau _v, \tau _I \right) = \left(\infty, \infty \right), \left(\infty, \tau \right), \text{and} \left(2\tau, \tau\right)$). 

\subsection{Neuron model with  $\tau _v = \infty$ and $\tau _I = \infty$}
When $\tau _I = \tau _v = \infty$, the membrane potential and firing time can be calculated as follows:
\begin{flalign}
v_i ^{(l)} (t) &= \sum _{j=1} ^{N^{(l-1)}} w_{ij}^{(l)} (t-t_j ^{(l-1)}) \theta(t - t_j ^{(l-1)}), \\
t_i^{(l)} &= \frac{V_\text{th} + \sum _{j\in \Gamma_i^{(l)}} w_{ij}^{(l)}t_j^{(l)}}{\sum _{j\in \Gamma_i^{(l)}} w_{ij}^{(l)}} \label{eq:timing_no_leak},\\
\hat{t}_i^{(l)} &= \frac{\hat{v} + \sum _{j\in \Gamma_i^{(l)}} w_{ij}^{(l)}t_j^{(l)}}{\sum _{j\in \Gamma_i^{(l)}} w_{ij}^{(l)}}, \\
\frac{t_i^{(l)}- \hat{t}_i^{(l)}}{V_\text{th} - \hat{v}} &= \frac{1}{\sum _{j\in\Gamma_i^{(l)}} w_{ij}}. \label{eq:t_diff_M_SSR}
\end{flalign}
Using these equations, the limit $(\hat{v}\rightarrow V_\text{th})$ can be calculated as follows:
\begin{flalign}
V_i^{(l)} &= \frac{1}{V_\text{th}-\hat{v}} \int _{\hat{t}_i^{(l)}} ^{t_i^{(l)}} dt v_i^{(l)}(t) \\
&= \frac{1}{V_\text{th}-\hat{v}} \int _{\hat{t}_i^{(l)}} ^{t_i^{(l)}} dt  \sum _{j\in \Gamma _i^{(l)}} w_{ij}^{(l)} \left( t - t_j^{(l-1)} \right)     \label{eq:V_v_insert}\\
&= \frac{1}{V_\text{th}-\hat{v}} \left[ \frac{1}{2}\sum _{j\in\Gamma_i^{(l)}} w_{ij}^{(l)} \left( t - t_j^{(l-1)} \right)^2 \right]_{\hat{t}_i^{(l)}}^{t_i^{(l)}}  \label{eq:V_3} \\
&= \frac{1}{2\left(V_\text{th}-\hat{v}\right)}  \sum _{j\in\Gamma_i^{(l)}} w_{ij}^{(l)} \left[\left( {\color{\fix}t_i^{(l)}} - t_j^{(l-1)} \right)^2 - \left( {\color{\fix}\hat{t}_i^{(l)}} - t_j^{(l-1)} \right)^2\right] \\
&= \frac{1}{2} \frac{{\color{\fix}t_i^{(l)}} - {\color{\fix}\hat{t}_i^{(l)}}}{V_\text{th}-\hat{v}} \sum _{j\in\Gamma_i^{(l)}} w_{ij}  \left( {\color{\fix}t_i^{(l)}} + {\color{\fix}\hat{t}_i^{(l)}} - 2 t_j^{(l-1)} \right)  \\
&\overset{\text{Eq.} (\ref{eq:t_diff_M_SSR})}{=} \frac{1}{\color{\fix} \sum _{j\in\Gamma_i^{(l)}} w_{ij}}\left[ {\color{\fix}t_i^{(l)}} \sum _j w_{ij} - \sum _j w_{ij}t_j^{(l-1)} \right]. \label{eq:V_limit} 
\end{flalign}
Note that the blue variables are related to the integral range, which is treated as a fixed value when calculating the gradient, as mentioned above.  
%例えば, にEq. (\ref{eq:timing_no_leak})をEq. (\ref{eq:V_limit})に代入することで$\sum _i \frac{1}{2\sum _j w_{ij}}$が得られる.
%この場合, 重みを大きくすればするほど損失が小さくなるが, 
%重みを大きくすれば, 発火は促進されることは明白であるため, 
%導入した損失項の求める働きと異なることがわかる.  
%一方で, 式(\ref{eq:V_limit})では, 重みは小さくなるほうが, 損失を小さくできる. 
%正の重みに関する入力タイミングは, 遅くなるほうが, 損失を小さくできる. 
%負の重みに関する入力タイミングは, 早く(小さく)するほうが, 損失を小さくできる. 
%このように, 直感にあう結果になる. 

\subsection{Neuron model with $\tau _v = \infty$ and $\tau _I = \tau$}
When $\tau _v = \infty$ and $\tau _I = \tau$, the membrane potential and firing time are given by \cite{Mostafa2018supervised}
\begin{flalign}
v_i^{(l)}(t) &= \tau \sum_{j=1}^{N^{(l-1)}} w_{ij}^{(l)}\theta(t-t_j^{(l-1)}) \left[ 1 - \exp \left(- \frac{t-t_j^{(l-1)}}{\tau}\right)\right], \\
t_i^{(l)} &= \tau \ln\left[\frac{\sum _{j\in\Gamma_i^{(l)}} w_{ij}^{(l)} \exp \left( \frac{t_j^{(l-1)}}{\tau} \right)}{\sum _{j\in\Gamma_i^{(l)}} w_{ij}^{(l)} - V_\text{th}\tau^{-1}}\right], \\
\hat{t}_i^{(l)} &= \tau \ln\left[\frac{\sum _{j\in\Gamma_i^{(l)}} w_{ij}^{(l)} \exp \left( \frac{t_j^{(l-1)}}{\tau} \right)}{\sum _{j\in\Gamma_i^{(l)}} w_{ij}^{(l)} - \hat{v}\tau^{-1}}\right].
\end{flalign}
We also obtain the following in $\hat{v}\rightarrow V_\text{th}$: 
\begin{flalign}
\frac{t_i^{(l)} - \hat{t}_i^{(l)}}{V_\text{th} - \hat{v}} &=\frac{\tau}{V_\text{th} - \hat{v}} \ln \frac{\sum _{j\in\Gamma_i^{(l)}} w_{ij}^{(l)} \exp \left( \frac{t_j^{(l)}}{\tau} \right)}{\sum _{j\in\Gamma_i^{(l)}} w_{ij}^{(l)} - V_\text{th}\tau^{-1}} - \frac{\tau}{V_\text{th} - \hat{v}} \ln \frac{\sum _{j\in\Gamma_i^{(l)}} w_{ij}^{(l)} \exp \left( \frac{t_j^{(l)}}{\tau} \right)}{\sum _{j\in\Gamma_i^{(l)}} w_{ij}^{(l)} - \hat{v}\tau^{-1}} \\
&= \frac{\tau}{V_\text{th} - \hat{v}} \ln \frac{\sum _{j\in\Gamma_i^{(l)}} w_{ij}^{(l)} - \hat{v}\tau^{-1}}{\sum _{j\in\Gamma_i^{(l)}} w_{ij}^{(l)} - V_\text{th}\tau^{-1}} \\
&= \frac{\tau}{V_\text{th} - \hat{v}} \ln  \left( 1 + \frac{\left(V_\text{th}-\hat{v} \right)\tau^{-1}}{\sum _{j\in\Gamma_i^{(l)}} w_{ij}^{(l)} - V_\text{th}\tau^{-1}} \right) \\
&\overset{\hat{v}\rightarrow V_\text{th}}{=} \frac{1}{\sum _{j\in\Gamma_i^{(l)}} w_{ij}^{(l)} - V_\text{th}\tau^{-1}}. \label{eq:t_diff_M_SSR_Mostafa}
\end{flalign}

Using these, the limit $(\hat{v}\rightarrow V_\text{th})$ can be calculated as follows:
\begin{flalign}
&\frac{\int dt v_i^{(l)}(t) }{V_\text{th}-\hat{v}} = \frac{\tau}{V_\text{th}-\hat{v}} \left[ \sum_{j\in\Gamma_i^{(l)}}  w_{ij}^{(l)}\left\{ t + \tau \exp \left(\frac{t_j^{(l-1)}}{\tau}\right) \exp \left(- \frac{t}{\tau}\right)\right\} \right]^{t_i^{(l)}}_{\hat{t}_i^{(l)}} \\
&= \frac{\tau}{V_\text{th}-\hat{v}}  \sum_{j\in\Gamma_i^{(l)}}  w_{ij}^{(l)}\left[ ({\color{\fix}{t_i^{(l)}}} - {\color{\fix}{\hat{t}_i^{(l)}}}) + \tau \exp \left(\frac{t_j^{(l-1)}}{\tau}\right) \left\{ \exp \left(- \frac{{\color{\fix}{t_i^{(l)}}}}{\tau}\right) - \exp \left(- \frac{{\color{\fix}{\hat{t}_i}^{(l)}}}{\tau}\right) \right\}\right]  \\
&= \frac{\tau({\color{\fix}{t_i^{(l)}}} - {\color{\fix}{\hat{t}_i^{(l)}}})}{V_\text{th}-\hat{v}} \sum_{j\in\Gamma_i^{(l)}}  w_{ij}^{(l)}\left[ 1 + \frac{\tau}{({\color{\fix}{t_i^{(l)}}} - {\color{\fix}{\hat{t}_i^{(l)}}})} \exp \left(\frac{t_j^{(l-1)}}{\tau}\right) \exp \left(- \frac{{\color{\fix}{t_i^{(l)}}}}{\tau}\right) \left\{ 1 - \exp \left( \frac{{\color{\fix}{t_i^{(l)}}} - {\color{\fix}{\hat{t}_i^{(l)}}}}{\tau}\right) \right\}\right]   \\
&\overset{\hat{v}\rightarrow V_\text{th}}{=} \tau {\frac{1 }{\textcolor{\fix}{\sum _{j\in\Gamma_i^{(l)}} w_{ij}^{(l)}} - V_\text{th}\tau ^{-1}}}  \sum_{j\in\Gamma_i^{(l)}}  w_{ij}^{(l)} \left\{ 1 - \exp \left(\frac{t_j^{(l)}}{\tau}\right) \exp \left(- \frac{{\color{\fix}{t_i^{(l)}}}}{\tau}\right)  \right\}  \\
&=\tau {\frac{1 }{\textcolor{\fix}{\sum _{j\in\Gamma_i^{(l)}} w_{ij}^{(l)}} - V_\text{th}\tau ^{-1}}} \left[ \left( \sum_{j\in\Gamma_i^{(l)}}  w_{ij}^{(l)} \right) - \exp \left(- \frac{{\color{\fix}{t_i^{(l)}}}}{\tau}\right) \sum_{j\in\Gamma_i^{(l)}}  w_{ij}^{(l)}  \exp \left(\frac{t_j^{(l-1)}}{\tau}\right)    \right].
\end{flalign}

\subsection{Alpha-synaptic neuron model with $\tau _v = 2\tau _I = 2\tau$} \label{ss:alpha_synaptic}
When $\tau _v = 2 \tau _I = 2 \tau$, the membrane potential is given by
\begin{flalign}
v_i^{(l)}(t) &= 2\tau \sum_{j\in\Gamma_i^{(l)}} w_{ij}^{(l)}\theta(t-t_j^{(l-1)}) \left[ \exp \left(-\frac{t-t_j^{(l-1)}}{2\tau }\right) - \exp \left(- \frac{t-t_j^{(l-1)}}{\tau }\right)\right].
\end{flalign}
From the firing condition $v_i^{(l)}\left(t_i^{(l)}\right)=V_\text{th}$, we obtain the following:
\begin{flalign}
 \exp \left(-\frac{t_i^{(l)}}{2\tau }\right) \sum_{j\in\Gamma_i^{(l)}} w_{ij}^{(l)} \exp \left(\frac{t_j^{(l-1)}}{2\tau }\right) - \exp \left(-\frac{t_i^{(l)}}{\tau }\right) \sum_{j\in\Gamma_i^{(l)}} w_{ij}^{(l)} \exp \left(\frac{t_j^{(l-1)}}{\tau }\right)&= \frac{V_\text{th}}{2\tau} \\
\left[\exp \left(-\frac{t_i^{(l)}}{2\tau }\right)\right]^2 \sum_{j\in\Gamma_i^{(l)}} w_{ij}^{(l)} \exp \left(\frac{t_j^{(l-1)}}{\tau }\right) - \exp \left(-\frac{t_i^{(l)}}{2\tau }\right) \sum_{j\in\Gamma_i^{(l)}} w_{ij}^{(l)} \exp \left(\frac{t_j^{(l-1)}}{2\tau }\right)  +   \frac{V_\text{th}}{2\tau} &=  0  \\
a_i \left[\exp \left(-\frac{t_i^{(l)}}{2\tau }\right)\right]^2  - b_i^{(l)} \exp \left(-\frac{t_i^{(l)}}{2\tau }\right)   +   \frac{V_\text{th}}{2\tau} &= 0, \label{eq:quadratic}
\end{flalign}
where we defined the following variables:
\begin{flalign}
a_i^{(l)}= \sum _{j\in\Gamma_i^{(l)}} w_{ij}^{(l)} \exp \left( \frac{t_j^{(l-1)}}{\tau} \right),~  b_i^{(l)}= \sum _{j\in\Gamma_i^{(l)}} w_{ij}^{(l)} \exp \left( \frac{t_j^{(l-1)}}{2\tau} \right).
\end{flalign}
From the formula to solve a quadratic function, the firing time $t_i^{(l)}$ can be calculated from Eq. (\ref{eq:quadratic}) as follows \cite{Goltz2021fast}: 
\begin{flalign}
\exp \left(-\frac{t_i^{(l)}}{2\tau }\right) &=  \frac{b_i^{(l)} + \sqrt{(b_i^{(l)})^2 - 2a_i^{(l)}\tau^{-1}V_\text{th}}}{2a_i^{(l)}} \\
t_i^{(l)} &= -2\tau \ln\left[\frac{b_i^{(l)} + \sqrt{(b_i^{(l)})^2 - 2a_i^{(l)}\tau^{-1}V_\text{th}}}{2a_i^{(l)}} \right], \\
%&= 2\tau \ln\left[\frac{2a_i^{(l)}}{b_i^{(l)} + \sqrt{(b_i^{(l)})^2 - 2a_i^{(l)}\tau^{-1}V_\text{th}}} \right] \label{eq:timing_alpha_1} \\
%\biggl(&= 2\tau \ln\left[\tau\frac{b_i^{(l)} - \sqrt{(b_i^{(l)})^2 - 2a_i^{(l)}\tau^{-1}V_\text{th}}}{V_\text{th}} \right]\biggr). \label{eq:timing_alpha_2}
\end{flalign}
where the other solution of the quadratic function is ignored because it indicates the time at which the membrane potential decreases from a value greater than $V_\text{th}$ to a smaller value \cite{Goltz2021fast}.
We also obtain the following:
\begin{flalign}
\frac{t_i^{(l)} - \hat{t}_i^{(l)}}{V_\text{th} - \hat{v}} &=  \frac{2\tau}{V_\text{th} - \hat{v}} \ln \frac{b_i^{(l)} + \sqrt{(b_i^{(l)})^2-2a_i^{(l)}\tau^{-1}\hat{v}}}{b_i^{(l)} + \sqrt{(b_i^{(l)})^2-2a_i^{(l)}\tau^{-1}V_\text{th}}} \\ 
&= \frac{2\tau}{V_\text{th} - \hat{v}} \ln \frac{b_i^{(l)} + \sqrt{(b_i^{(l)})^2-2a_i^{(l)}\tau^{-1}V_\text{th} + 2a_i^{(l)}\tau^{-1}\left(V_\text{th} - \hat{v}\right)}}{b_i^{(l)} + \sqrt{(b_i^{(l)})^2-2a_i^{(l)}\tau^{-1}V_\text{th}}} \\
&\overset{\hat{v}\rightarrow V_\text{th}}{=} 2 \tau \frac{a_i^{(l)} \tau ^{-1}} {\left(b_i^{(l)} + \sqrt{(b_i^{(l)})^2-2a_i^{(l)}\tau^{-1}V_\text{th}}\right) \sqrt{(b_i^{(l)})^2-2a_i^{(l)}\tau^{-1}V_\text{th}}} \label{eq:t_diff_alpha_taylor} \\
&= \alpha_i^{(l)}.   \label{eq:t_diff_M_SSR_alpha}
\end{flalign}
We define the following variable: 
\begin{flalign}
\alpha_i^{(l)} = \frac{2a_i}{\left(b_i^{(l)} + \sqrt{(b_i^{(l)})^2-2a_i^{(l)}\tau^{-1}V_\text{th}}\right)\left(\sqrt{(b_i^{(l)})^2 - 2a_i^{(l)}\tau^{-1}V_\text{th}}\right)}.
\end{flalign}
Using these, the limit $(\hat{v}\rightarrow V_\text{th})$ can be calculated as follows:
\begin{flalign}
&\frac{\int v_i^{(l)}(t) dt}{V_\text{th}-\hat{v}} = \frac{2\tau}{V_\text{th}-\hat{v}} \sum_{j\in\Gamma_i^{(l)}}  \left[w_{ij} \left\{ -2\tau \exp \left(-\frac{t-t_j^{(l)}}{2\tau}\right) + \tau \exp \left(- \frac{t-t_j^{(l-1)}}{\tau}\right)\right\} \right]^{t_i^{(l)}}_{{\hat{t}_i^{(l)}}}\\
&= \frac{2\tau^2}{V_\text{th}-\hat{v}} \sum_{j\in\Gamma_i^{(l)}}  w_{ij}^{(l)} \left[  -2\exp \left(\frac{t_j^{(l-1)}}{2\tau}\right) \exp \left(-\frac{t}{2\tau}\right) + \exp \left(\frac{t_j^{(l-1)}}{\tau}\right) \exp \left(- \frac{t}{\tau}\right) \right]^{t_i^{(l)}}_{{\hat{t}_i^{(l)}}}\\
&= \frac{2\tau^2}{V_\text{th}-\hat{v}} \sum_{j\in\Gamma_i^{(l)}}  w_{ij}^{(l)}  \Bigg[-2 \exp \left(\frac{t_j^{(l-1)}}{2\tau}\right)  \left\{\exp \left(-\frac{\color{\fix}{t_i^{(l)}}}{2\tau}\right) - \exp \left(-\frac{{\color{\fix}{\hat{t}_i^{(l)}}}}{2\tau}\right) \right\}  \nonumber \\
&~~~~~~~~~~~~~~~+ \exp \left(\frac{t_j^{(l-1)}}{\tau}\right) \left\{ \exp \left(- \frac{{\color{\fix}{t_i^{(l)}}}}{\tau}\right) - \exp \left(- \frac{{\color{\fix}{\hat{t}_i^{(l)}}}}{\tau}\right)\right\} \Bigg]\\
&= \frac{2\tau^2 \left({\color{\fix}{t_i^{(l)}}} - {\color{\fix}{\hat{t}_i^{(l)}}} \right)}{V_\text{th}-\hat{v}} \sum_{j\in\Gamma_i^{(l)}}  \frac{w_{ij}^{(l)}}{{\color{\fix}{t_i^{(l)}}} - {\color{\fix}{\hat{t}_i^{(l)}}} } \Bigg[-2 \exp \left(\frac{t_j^{(l-1)}}{2\tau}\right) \exp \left(-\frac{{\color{\fix}{t_i^{(l)}}}}{2\tau}\right) \left\{1 - \exp \left(\frac{{\color{\fix}{t_i^{(l)}}} - {\color{\fix}{\hat{t}_i^{(l)}}}}{2\tau}\right) \right\} \nonumber \\
&~~~~~~~~~~~~~~~+ \exp \left(\frac{t_j^{(l-1)}}{\tau}\right) \exp \left(- \frac{{\color{\fix}{t_i^{(l)}}}}{\tau}\right) \left\{ 1 - \exp \left(\frac{{\color{\fix}{t_i^{(l)}}} - {\color{\fix}{\hat{t}_i^{(l)}}}}{\tau}\right) \right\} \Bigg]\\
&\overset{\hat{v} \rightarrow V_\text{th}}{=} 2\tau {\color{\fix}{\alpha _i^{(l)}}} \left[ \exp \left(-\frac{{\color{\fix}{t_i^{(l)}}}}{2\tau}\right) \sum_{j\in\Gamma_i^{(l)}}  w_{ij}^{(l)}   \exp \left(\frac{t_j^{(l-1)}}{2\tau}\right)  - \exp \left(-\frac{{\color{\fix}{t_i^{(l)}}}}{\tau}\right) \sum_{j\in\Gamma_i^{(l)}}  w_{ij}^{(l)}   \exp \left(\frac{t_j^{(l-1)}}{\tau}\right) \right] \\
&=  2\tau{\color{\fix}{\alpha _i^{(l)}}} \left[ \exp \left(-\frac{{\color{\fix}{t_i^{(l)}}}}{2\tau}\right) b_i^{(l)}  - \exp \left(-\frac{{\color{\fix}{t_i^{(l)}}}}{\tau}\right) a_i^{(l)} \right].
\end{flalign}

\section{Numerical convergence analysis} \label{ss:convergence}

M-SSR was derived by taking the limit of $\hat{v}\rightarrow V_\text{th}$ for the integral-form regularization (Eq. \ref{eq:Vmem_integration}). 
In this section, we confirm that the gradient of the integral-form regularization (Eq. \ref{eq:Vmem_integration}) converges to that of M-SSR (Eq. \ref{eq:M_SSR}) when $\hat{v}$ approaches $V_\text{th}$ through numerical simulation.   
The Iris dataset \cite{scikit-learn} was used as input data. 
The Iris dataset consists of 150 instances of three-class data with four features. 
To process the data with an SNN, each feature was normalized to $[0,1]$  and each element of the vector ($x_i$) was transformed to the time $t_i^{(0)}=\tau_\text{in} x_i~(i=0,1,2,3)$ of the input layer spikes, where $\tau_\text{in}=5$. 
We also introduced a bias spike $t_4^{(0)}=0$. 
This input spike was input to an SNN with two hidden layers (5-10-10-3). 
The cost function (Eq. (\ref{eq:cost_function})) was active only for $V$, with $\gamma_2=1$.
The error for the gradient of the weights for each layer was defined as follows:
\begin{flalign}
Error^{(l)}(x) =  \frac{1}{N^{(l)}N^{(l-1)}}\sum _{i=1}^{N^{(l)}}\sum _{j=1}^{N^{(l-1)}} \frac{\left|\frac{\partial C(\hat{v}=x)}{\partial w_{ij}^{(l)}} - \frac{\partial C(\hat{v}\rightarrow V_\text{th})}{\partial w_{ij}^{(l)}}\right|}{\left|\frac{\partial C(\hat{v}=x)}{\partial w_{ij}^{(l)}}\right|}.
\end{flalign}
The integral-form regularization (Eq. (\ref{eq:Vmem_integration})) was computed by setting $T=t^\text{ref}$ and dividing the integral by the time step width $\Delta t=\frac{T}{N_\text{steps}}$.

\begin{figure*}%[h]
\begin{center}
\includegraphics[clip,width=16cm]{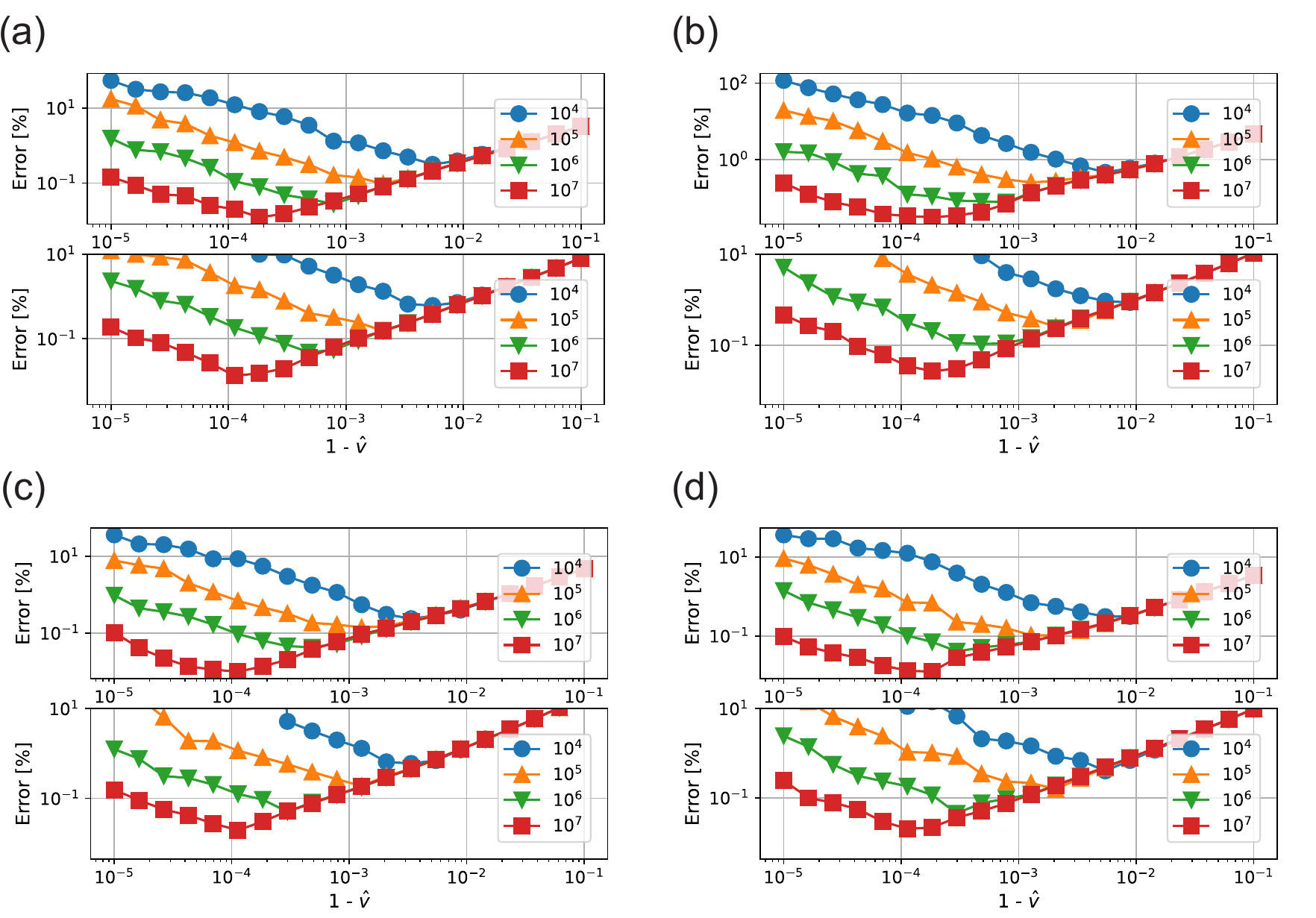}
\caption{{\bf Numerical conversion analysis.}  The gradient errors in the SNNs (5-10-10-3) for various values of $\hat{v}$ and $N_\text{steps} \in \{10^4, 10^5, 10^6, 10^7\}$ are plotted. 
The results are shown for the following neuron models: (a) $\tau_v = \tau_I = \infty$, (b) $\tau_v=\infty$, $\tau_I = 5$, (c) $\tau_v = 2 \tau_I = 10$, and (d) $\tau_v = 2 \tau_I = 20$.
In each subfigure, the upper panel shows the gradient error relating to the first hidden layer and the lower panel shows the gradient error relating to the second hidden layer. 
The gradient error was calculated by averaging over the 150 data. 
We set $t^\text{ref}=10$. 
}
\label{fig:convergence}
\end{center}
\end{figure*}

Figure \ref{fig:convergence} shows the results of the gradient error for various neuron models.
The gradient error was calculated by averaging over the 150 data. 
For all neuron models and $N_\text{steps}$, we can confirm that the errors decrease as $\hat{v}$ is brought closer to $V_\text{th}$, and then the errors tend to increase. 
This increase in error is due to the fact that if $\hat{v}$ is too close to $V_\text{th}$, the integration range becomes extremely narrow, making it difficult to evaluate the integral.  
As $N_\text{steps}$ is increased, it is found that the integrals can be evaluated more precisely up to the point where $\hat{v}$ is closer to $V_\text{th}$. 
As $N_\text{steps}$ is increased, the minimum value of the gradient error decreases uniformly, and it is found to decrease to about $0.1\%$ for $N_\text{steps} = 10^7$.  

\section{Effects of $\xi$} \label{ss:magnification}

\begin{figure*}%[h]
\begin{center}
\includegraphics[clip,width=\textwidth]{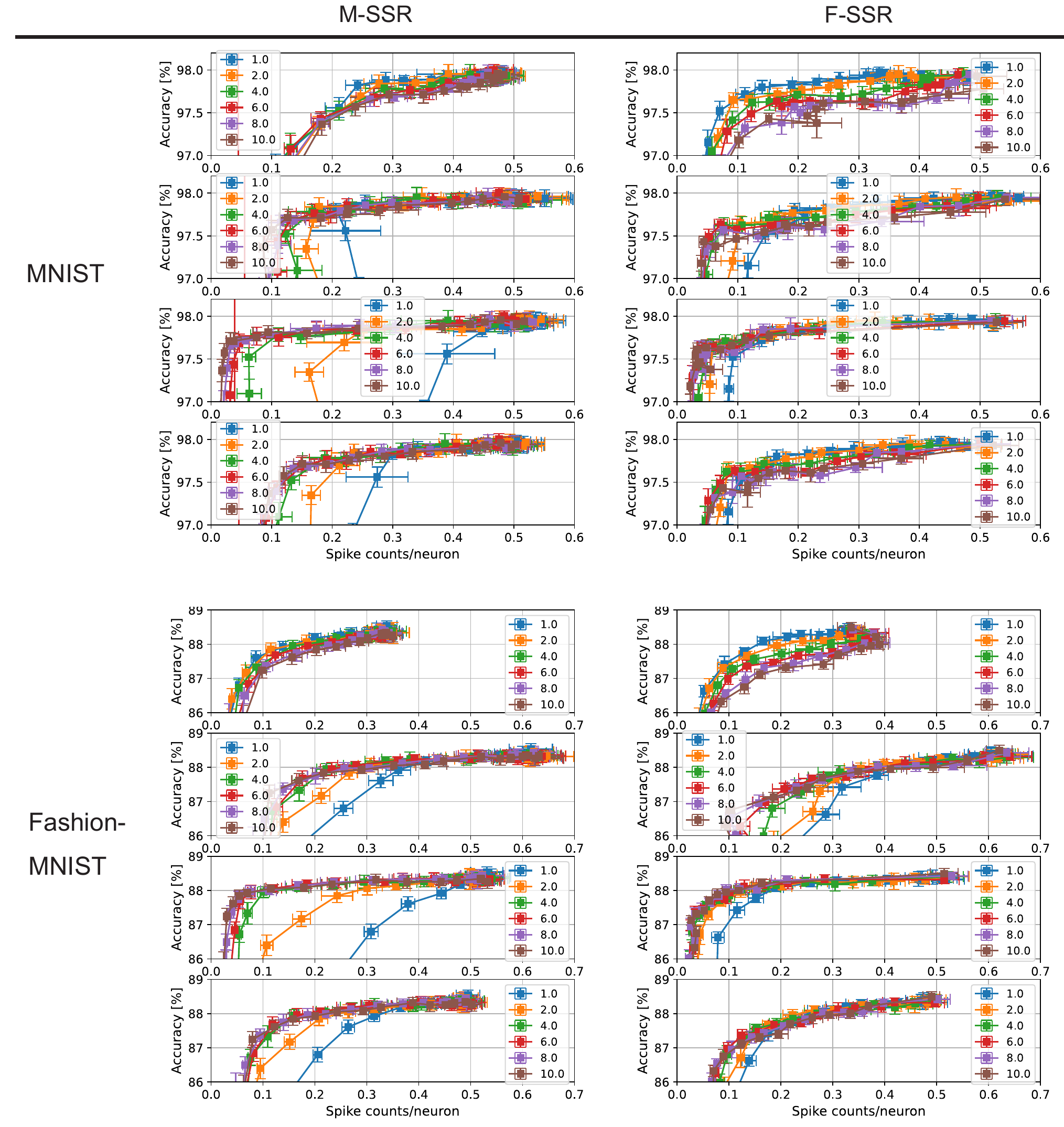}
\caption{{\bf Effect of $\xi$ on the sparsity--accuracy tradeoff.} 
The sparsity--accuracy tradeoffs are plotted for SNNs (784-400-400-400-10) with M-SSR and F-SSR for various values of $\xi$. 
For each dataset, the first three panels present the sparsity--accuracy tradeoff for the first, second, and third hidden layers, from the top. The bottom panel presents the sparsity--accuracy tradeoff for the sparsity averaged over the three hidden layers.
We used the following hyperparameters: $t^\text{ref}=9,~\gamma_1=10^{-4}, \eta=10^{-4}$, and $\tau_\text{soft}=0.9$. 
}
\label{fig:magnify}
\end{center}
\end{figure*}

In the proposed SSR methods (Eqs. (\ref{eq:Vmem_loss}) and (\ref{eq:F_loss})), the regularization loss is larger in the latter layers as the coefficient $\xi$ increases, resulting in sparse firing characteristics in the latter layers. 
Figure \ref{fig:magnify} shows the effect of the value of $\xi$ on the sparsity--accuracy tradeoff with M-SSR (Eq. (\ref{eq:M_SSR})) and F-SSR (Eq. (\ref{eq:F_SSR})). 
We trained the SNNs with three hidden layers (784-400-400-400-10), and the results are shown for the MNIST and Fashion-MNIST datasets. 
For each dataset, the top three panels show the sparsity--accuracy tradeoffs for the first, second, and third hidden layers, and the bottom panel shows the average results for all hidden layers.  
In the case of M-SSR, when $p_\text{layer}=1$, the latter layer is not sparse, but when $p_\text{layer}=6$, both layers are sparse. 
By contrast, in the case of F-SSR, even when $p_\text{layer}=1$, the latter layers are relatively sparse. 
This difference can be attributed to the fact that in M-SSR, the regularization error propagates backward to the previous layer, whereas in F-SSR, the regularization error is local and does not propagate backward.

\section{Raw plots of sparsity--accuracy tradeoff for CNN-SNNs} \label{ss:raw_plot}

\begin{figure*}%[h]
\begin{center}
\includegraphics[clip,width=13cm]{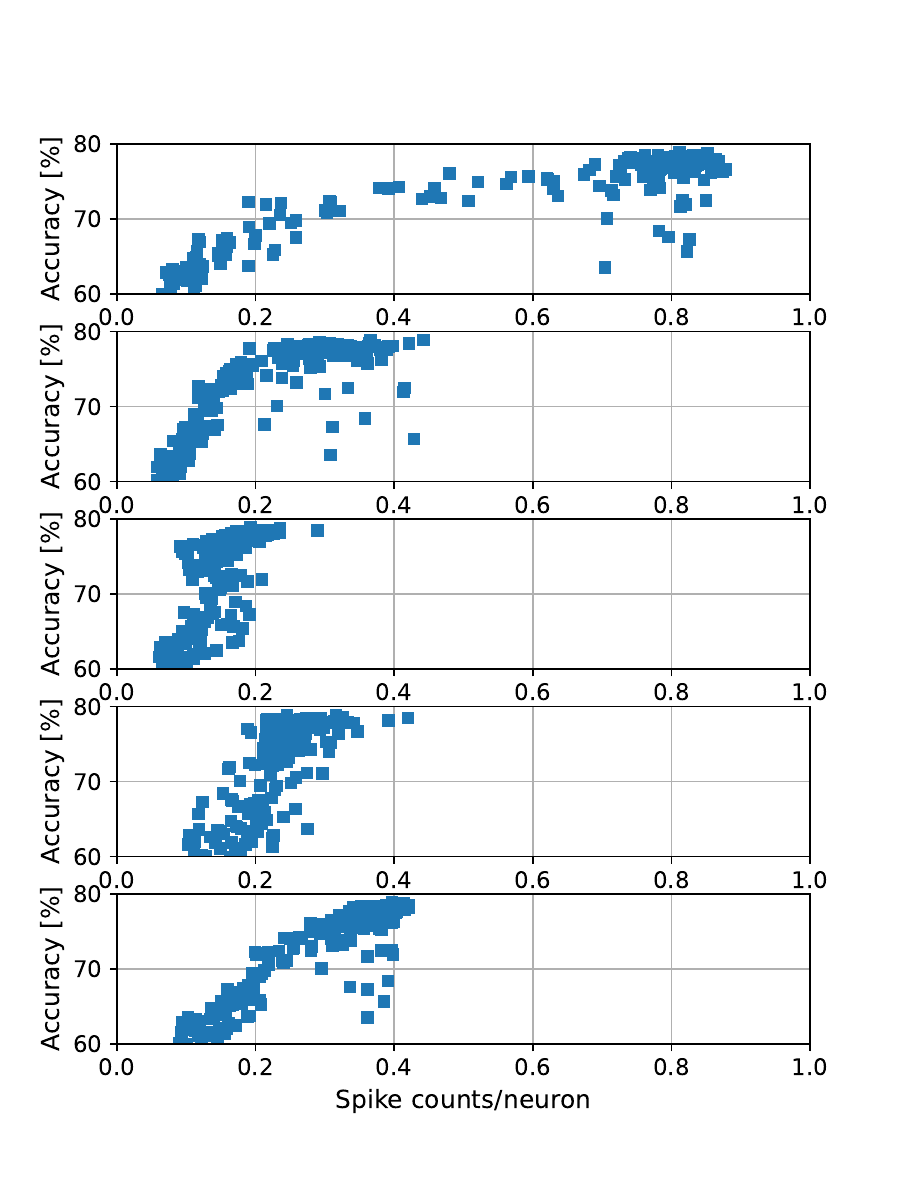}
\caption{{\bf Raw plot of the sparsity--accuracy tradeoff on the CIFAR-10 dataset.} 
}
\label{fig:VGG_raw}
\end{center}
\end{figure*}

Figure \ref{fig:VGG_raw} shows the sparsity--accuracy tradeoffs for SNNs trained on the CIFAR-10 dataset. 
The regularization strength of F-SSR $\gamma_2$ ranged from $0$ to $-10^{-6}$. 
The training was performed with 10 different initial weights for each value of $\gamma_2$. 
All 190 data points obtained are shown in the figure.

\bibliographystyle{unsrt}
\bibliography{myBib}

\end{document}